\documentclass[12pt]{article}
\usepackage{graphicx}
\usepackage{amsmath, amssymb, mathtools, bbm}
\usepackage{amsthm}
\usepackage{bm}
\usepackage{mathpazo}
\usepackage{booktabs, multirow}
\usepackage{natbib}
\usepackage[colorlinks=true, citecolor=blue, linkcolor=blue, urlcolor=blue]{hyperref}
\usepackage{microtype}
\usepackage{setspace}
\usepackage{caption}
\usepackage[letterpaper, left=1.25in, right=1.25in, top=1.2in, bottom=1.2in]{geometry}
\usepackage{subcaption}
\usepackage{enumitem}
\usepackage{subcaption}
\usepackage{placeins}

\newcommand{\indep}{\!\perp\!\!\!\perp}

\def\*#1{\mathbf{#1}}

\newcommand{\cF}{\mathcal{F}}
\newcommand{\cG}{\mathcal{G}}
\newcommand{\cR}{\mathcal{R}}
\newcommand{\cS}{\mathcal{S}}

\newcommand{\blind}{0}
\newcommand{\tit}{Using Embedding Models to Improve Probabilistic Race Prediction}

\begin{document}
\date{\today}

\title{{\bf\tit}\thanks{We thank Bruce Willsie of L2, Inc. for his generosity in sharing the voter files used in this paper.}}

\if0\blind

\author{Noah Dasanaike\thanks{PhD Candidate, Department of Government, Harvard University.} \and Kosuke Imai\thanks{Professor, Department of Government and
      Department of Statistics, Harvard University.  1737 Cambridge
      Street, Institute for Quantitative Social Science, Cambridge MA
      02138.  Email: \href{mailto:imai@harvard.edu}{imai@harvard.edu}
      URL:
      \href{https://imai.fas.harvard.edu}{https://imai.fas.harvard.edu}}}

\fi

\maketitle

\begin{abstract}
Estimating racial disparity requires individual-level race data, which are often unavailable due to the sensitivity of collecting such information.
To address this problem, many researchers utilize Bayesian Improved Surname Geocoding (BISG), which has critically relied on Census surname data.
Unfortunately, these data capture race-surname relationships only for common surnames, omitting approximately 10\% of the US population.
We show that predictive performance degrades substantially for individuals with such omitted, uncommon surnames because standard BISG implementation relies on an uninformative generic prior in these cases.
To address this limitation, we propose embedding-powered BISG (eBISG), which uses pre-trained text embeddings to represent names as dense vectors and trains neural networks on 2020 Census surname and first-name data to estimate race probabilities for names not covered in the Census.
We compare five approaches: standard BISG using only surnames, BIFSG incorporating first name probabilities, surname embedding for unlisted names, surname and first name embedding combining both, and a full-name embedding trained on voter file data from Southern states that captures interactions between name components.
We show that each successive eBISG approach improves race prediction, with the full-name embedding yielding the largest gains, particularly for Hispanic and Asian voters whose surnames are absent from the Census surname-by-race frequency table.
\end{abstract}

\newpage
\section{Introduction}

Measuring racial disparities across diverse social, political, and economic domains requires knowing the race of individuals in administrative records \citep[e.g.,][]{adjayegbewonyo2014bisg,fraga2018turnout,greenwald2024regulatory}. Yet, individual race information is often not available in these data owing to legal restrictions, privacy concerns, or the perceived sensitivity of collecting such information \citep{fiscella2006geocoding}.

The most widely adopted solution to this problem is Bayesian Improved Surname Geocoding (BISG). Using Bayes' rule, BISG combines a surname-based racial prior $P(R \mid S)$ drawn from the Census Bureau's surname-by-race frequency table with geographic racial composition $P(R \mid G)$ to produce individual-level race probabilities $P(R \mid G, S)$ under the assumption of conditional independence between surname and geography given race $S \indep G \mid R$ \citep{elliott2009using,imaikhanna2016,imai2022fbisg}. 

Since its introduction, BISG has become the standard approach for racial imputation across the social and health sciences. The Consumer Financial Protection Bureau adopted BISG for fair lending enforcement \citep{cfpb2014proxy}, health researchers use it to track disparities in vulnerable populations \citep{adjayegbewonyo2014bisg}, and political scientists rely on it to study various political outcomes with respect to race \citep{fraga2018turnout}. Widely used software implementations make BISG and its extensions accessible to applied researchers \citep{wru}. \citet{greenwald2024regulatory} document consequences of BISG prediction errors for fair lending enforcement.

Most applications of BISG rely critically on Census name data. The latest release, published in April 2026, provides race distributions for 156,620 surnames and 53,616 first names occurring at least 100 times in the 2020 Census. However, racial distributions for common first names were not publicly available prior to this release. As a result, typical applications have relied solely on surname data. Individuals whose surnames do not appear in the Census surname-by-race frequency table (approximately 10\% of the population) are assigned a generic prior equal to the national racial distribution, which conveys no information about their likely race.

Using North Carolina and Florida voter files where self-reported race provides ground truth, we show that the predictive performance of BISG substantially degrades for these voters, with precision-recall performance falling sharply relative to voters with matched surnames.  These unmatched voters are not a random cross-section of the electorate: 16 to 18 percent of Asian voters and 10 to 12 percent of Hispanic voters have unmatched surnames, compared with 6 to 11 percent of white voters.

To address this limitation, we propose a class of methods called embedding-powered BISG (eBISG) that replaces the uninformative generic prior for unmatched surnames with informative, name-specific race probabilities derived from embeddings, while leaving Census-based priors and the BISG framework otherwise unchanged. Our approach uses open-source pre-trained text embedding models to represent names as dense vectors in a high-dimensional space, such that morphologically and linguistically similar names receive similar representations. In our implementation, we use E5-Large \citep{wang2024e5}, though alternative embedding models could also be used.

We evaluate the performance of five approaches, including three of our own, that progressively incorporate more name information. First, we consider two existing methods. Standard BISG uses only surname probabilities from the Census. BIFSG (Bayesian Improved First name and Surname Geocoding) supplements this with first name probabilities, treating surname, first name, and geolocation as conditionally independent of one another given race, i.e., $P(S, F, G \mid R) = P(S \mid R) P(F \mid R) P(G \mid R)$ \citep{voicu2018firstname}. As mentioned earlier, the latest release of the 2020 Census name data includes the information about frequently occurring first names, which makes the application of BIFSG straightforward.

Second, we propose three versions of eBISG. Our surname embedding approach replaces the uninformative generic prior for unmatched surnames with an embedding-based prediction trained on the Census surname-by-race frequency table. Our surname and first name embedding approach extends this by training a separate embedding model on the Census first name list, incorporating both as Bayesian factors for voters with unlisted names. Our full-name embedding approach, trained on voter files from Southern states, concatenates first, middle, and last name into a single representation. This allows the model to capture interactions between name components that are otherwise precluded by the conditional independence assumption. 

Because all three eBISG methods modify only the prior assigned to unmatched names, predictions are left unchanged for the roughly 90 percent of voters whose surnames appear in the Census surname-by-race frequency table. Furthermore, the proposed eBISG methodology seamlessly integrates with any downstream analysis that takes BISG posteriors as inputs, including the BIRDiE estimator introduced by \citet{mccartan2025birdie}.

We empirically evaluate all of these approaches using voter registration data from North Carolina and Florida, training the full-name model on voter files from four other Southern states to ensure a clean separation between training and evaluation samples while requiring generalization. We find that the approach based on surname embeddings yields consistent improvements over standard BISG for voters with unmatched names, while the method, which leverages full-name embeddings, produces substantially larger gains across all racial categories in both states. The full-name approach also reduces the systematic, income-correlated bias identified by \citet{argylebarber2024}. Where standard BISG overpredicts white share in wealthy Census tracts and underpredicts minority share across the income distribution, the full-name embeddings approach reduces these biases toward zero.

\subsection*{Related Literature}

A growing literature has sought to extend beyond the Census surname-by-race frequency table by training machine learning models on labeled data to classify individuals by race from their names. \citet{sood2018predicting}, for instance, show that character-level recurrent neural networks can achieve high accuracy on Florida voter data, while subsequent work has leveraged fine-tuned transformers \citep{parasurama2021racebert} and interpretable feature-based models that use name popularity patterns rather than character sequences \citep{jain2022importance}. \citet{ye2017nationality} and \citet{ye2019secret} learn name embeddings from communication patterns and social media data rather than from labeled racial categories. 

These methods classify each individual to a single racial category and are proposed as standalone alternatives to BISG.  However, probabilistic predictions play an essential role in conducting an unbiased downstream analysis.  \citet{mccartan2025birdie} show that the racial disparity estimates based on race classification are severely biased due to the failure to properly account for classification error.  In contrast to the existing approaches mentioned above, our eBISG method is designed to improve probabilistic race prediction and can be easily integrated into the existing analysis pipeline, e.g., the standard BISG procedure or its extension followed by the downstream analysis methods such as BIRDiE of \citet{mccartan2025birdie}.

A separate body of work has sought to incorporate first names into the BISG framework. \citet{voicu2018firstname} introduced BIFSG, which adds first name information as a separate Bayesian component alongside surname and geography, and demonstrated downstream improvements in racial predictions for mortgage lending applications. \citet{imai2022fbisg} extended this approach with a fully Bayesian model (fBISG) that accounts for possible measurement error in Census surname data and supplements the Census surname-by-race frequency table with name dictionaries drawn from the voter files of six Southern states, covering 136 thousand first names, 125 thousand middle names, and 338 thousand surnames \citep{rosenman2023names}. These approaches treat first, middle, and last names as conditionally independent signals, each of which contributes a separate Bayesian factor. Our eBISG method instead represents the full name as a single vector, allowing the model to capture interactions between given names and surnames that are otherwise precluded by the conditional independence assumption.

Recent work has also considered information beyond text for inferring racial group membership. \citet{leevelez2025} combine BISG with a convolutional neural network trained on publicly available photographs to predict the race of elected officials, while \citet{deluca2024power} evaluate machine-learning-modified BISG for redistricting applications. The embedding framework that we propose is complementary to these approaches, and a pre-trained multimodal model could in principle embed images alongside names, thus extending the same logic to visual information. 

The remainder of the paper is organized as follows. Section~2 documents the unmatched surname problem and its consequences for BISG performance. Section~3 describes our eBISG methodology. Section~4 presents the empirical evaluation, spanning individual-level race prediction, prediction quality across the income distribution, and calibration. Section~5 concludes.

\section{BISG Race Prediction with Unmatched Names}

\begin{table}[!t]
\centering
\begin{tabular}{l rr r}
\toprule
 & \multicolumn{2}{c}{Individuals with listed name (\%)} & \\
\cmidrule(lr){2-3}
Race & Surname & First name & US population (\%) \\
\midrule
White    & 60.0 & 59.8 & 57.8 \\
Black    & 11.8 & 11.8 & 12.1 \\
Hispanic & 18.0 & 18.2 & 18.7 \\
Asian    &  6.2 &  6.2 &  5.9 \\
AIAN     &  0.7 &  0.7 &  0.7 \\
Two or more &  3.3 &  3.3 &  4.1 \\
\midrule
Names listed & 156,620 & 53,616 & \\
Individuals covered & 298,870,618 & 302,029,327 & 331,449,281 \\
Coverage (\%) & 90.2 & 91.1 & \\
\bottomrule
\end{tabular}
\caption{Racial composition of individuals whose surnames or first names appear in the 2020 Census name files. The surname file lists 156,620 names occurring at least 100 times; the first name file lists 53,616 names. Each covers approximately 90\% of the US population. The fourth column reports 2020 Census population shares for comparison.}
\label{tab:census_names}
\end{table}

The 2020 Census name files, released in April 2026, tabulate the racial distribution of 156,620 surnames and 53,616 first names, each occurring at least 100 times in the decennial Census. As summarized in Table~\ref{tab:census_names}, the surname file accounts for roughly 90 percent of the US population, and the first name file for 91 percent. The racial composition of individuals with listed names closely tracks the overall US population. 

\begin{table}[t]
\centering
\begin{tabular}{l rr rr}
\toprule
 & \multicolumn{2}{c}{Surname matched (\%)} & \multicolumn{2}{c}{Surname unmatched (\%)} \\
\cmidrule(lr){2-3} \cmidrule(lr){4-5}
& \multicolumn{2}{c}{First name} & \multicolumn{2}{c}{First name} \\
Race & matched & unmatched & matched & unmatched \\
\midrule
White    & 66.0 & 1.0 & 49.7 & 1.4 \\
Black    & 15.9 & 2.5 &  8.5 & 2.4 \\
Hispanic &  9.5 & 0.6 & 27.9 & 2.8 \\
Asian    &  1.7 & 0.4 &  2.4 & 1.0 \\
AIAN     &  0.5 & 0.0 &  0.3 & 0.0 \\
Other    &  1.6 & 0.2 &  3.0 & 0.6 \\
\midrule
\% of group & 95.3 & 4.7 & 91.8 & 8.2 \\
\midrule
$N$ (surname group) & \multicolumn{2}{r}{1,753,274 (87.7\%)} & \multicolumn{2}{r}{246,721 (12.3\%)} \\
\bottomrule
\end{tabular}
\caption{Racial composition of voters by surname and first name match status in the 2020 Census name files, pooling North Carolina and Florida ($N = 1{,}999{,}995$). Percentages sum to 100 within each surname group (matched or unmatched) across both first name columns. State-by-state results appear in Table~\ref{tab:unmatched_bystate} of Appendix Section~\ref{sec:bystate}.}
\label{tab:unmatched}
\end{table}

However, when compared to the Florida and North Carolina voter files that we use for evaluation, we find that the 100-occurrence threshold disproportionately excludes names that are common among immigrant-origin communities. In particular, among individuals with an unmatched surname, Hispanic and Asian voters are overrepresented relative to their share of the matched population. 
Table~\ref{tab:unmatched} cross-tabulates surname and first name match status.\footnote{The voter files used in this study were sourced from L2, Inc., a leading national non-partisan firm and the oldest organization in the United States that supplies voter data and related technology to candidates, political parties, pollsters, and consultants for use in campaigns. The replication data for this paper can be purchased from L2, Inc.} Among voters with matched surnames, 95 percent also have a first name that appears in the Census first name file, leaving relatively few voters for whom only surname information is available. The more consequential gap is among voters with unmatched surnames, who account for approximately 12 percent of the pooled sample. These voters are heavily represented among certain minority groups. Hispanic voters, for instance, comprise 28 percent of all voters whose surnames are unmatched and first names matched, compared to 10 percent of the surname-matched population. Asian voters are similarly overrepresented (2.4 versus 1.7 percent). This reflects the reliance of the Census surname-by-race frequency table on a threshold that filters out surnames which are common in immigrant-origin communities but rare in the broader population.

\begin{figure}[t]
\centering
\includegraphics[width=\textwidth]{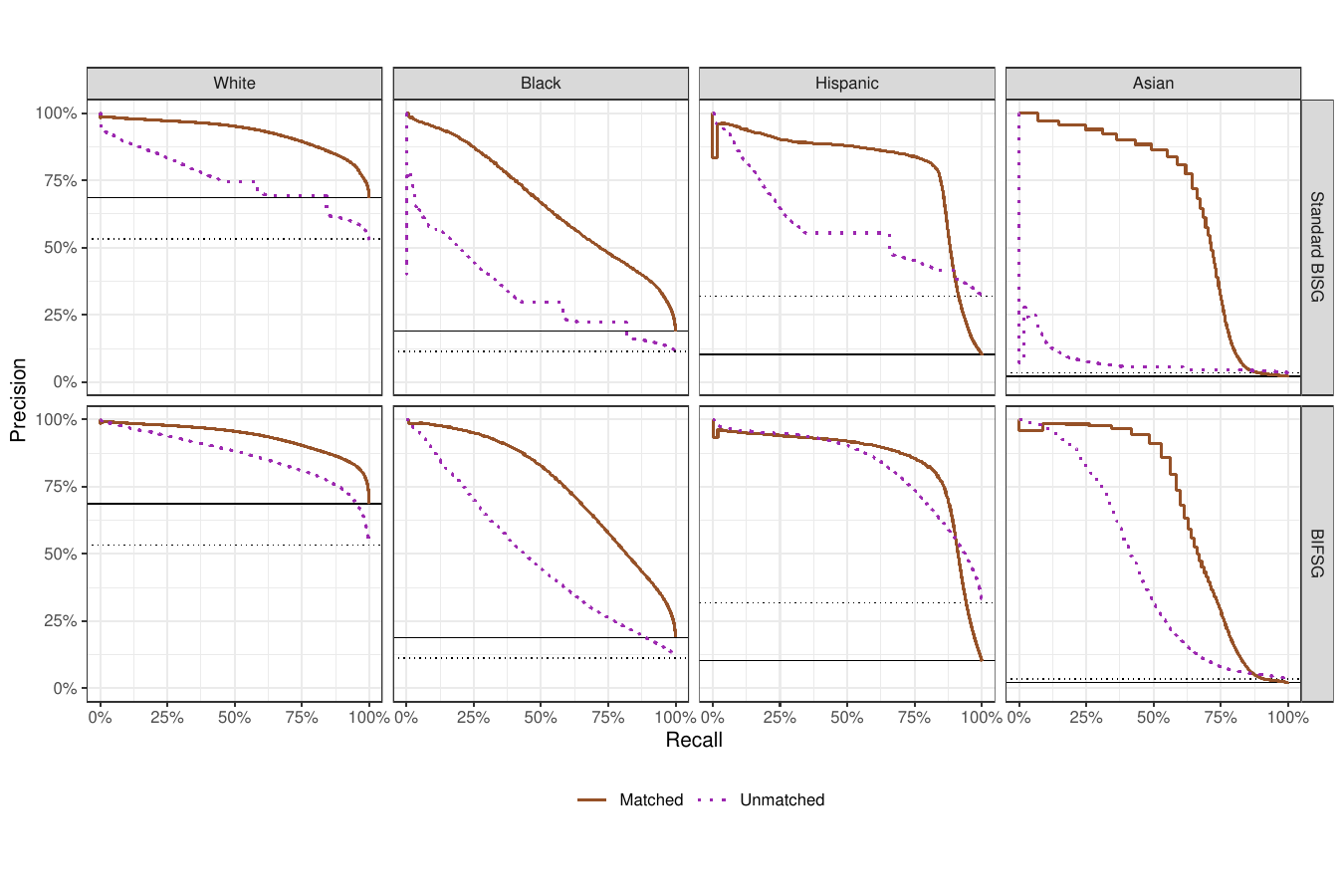}
\vspace{-.5in}
\caption{Precision-recall curves comparing voters whose surnames appear in the Census surname-by-race frequency table (matched, solid) with those whose surnames do not (unmatched, dotted), pooling North Carolina and Florida. Top row: standard BISG. Bottom row: BIFSG. The dotted horizontal line indicates the baseline classifier.}
\label{fig:pr_matched_unmatched}
\end{figure}

Figure~\ref{fig:pr_matched_unmatched} illustrates the consequence of unmatched surnames for probabilistic prediction. We plot precision-recall curves separately for voters with matched and unmatched surnames, pooling across North Carolina and Florida. The top row shows the performance of standard BISG using only the surname probabilities. As expected, precision drops sharply for unmatched voters across all four racial categories. For Asian and Hispanic voters with unmatched surnames, BISG barely exceeds the performance of the baseline classifier. The step-like pattern visible for Asian voters reflects the uninformative, generic prior. Because all unmatched voters receive the same surname prior $P(R)$, their BISG posteriors are determined entirely by tract-level geography. This concentrates the predicted probabilities in a narrow range, and produces a sharp transition in the precision-recall curve when the classification threshold crosses this range. 

The bottom row presents BIFSG. First names reduce the gap substantially, confirming that first names carry independent racial information. However, with the exception of Hispanics, a substantial difference persists between matched and unmatched surnames, because the latter still receive an uninformative prior that first names alone are unable to fully compensate for. 

\section{The Proposed eBISG Methodology}

In this section, we introduce the proposed eBISG methodology.  We begin by briefly reviewing the standard BISG and then show how we estimate the racial probabilities for surnames using embeddings. Finally, we extend our eBISG methodology by leveraging the embeddings for full names.

\subsection{Review of BISG and BIFSG}

Suppose that we have a random sample of $n$ individuals from a population.  We consider the problem of predicting the race of each individual $R \in \cR$ using the location of their residence $G \in \cG$ and surname $S \in \cS$ where $\cR$, $\cG$, and $\cS$ represent the sets of racial categories, geolocations, and surnames, respectively.  In our applications, $\cR$ represents the following six Census racial categories --- White, Black, Hispanic, Asian (including Native Hawaiian and Other Pacific Islander), American Indian or Alaska Native (AIAN), and Others.

BISG relies on two sources of information provided by the Census Bureau. The first is the surname-by-race frequency table, which reports the counts of individuals $N_s$ for each surname $s \in \cS_{M} \subset \cS$, where $\cS_M$ denotes the set of surnames included in the table. In the 2020 Census, this information was released for a total of 156,620 common surnames that occur at least 100 times. From this table, we can compute the conditional distribution of race given each common surname, $P(R \mid S)$.  Note that this information is not available for uncommon names $\cS_{U} = \cS \setminus \cS_{M}$.  In such cases, a typical implementation of BISG uses an overall proportion of racial group, i.e., $P(R)$.

Second, BISG uses the Census counts of each racial category within each geographical unit, which provides the conditional probability of geography given race, i.e., $P(G \mid R)$.  The key assumption of BISG is the following conditional independence between surname and geography given race, i.e., 
$$
S \indep G \mid R.
$$
This assumption may be violated if surnames remain informative of geolocation even after conditioning on race. For example, distinct ethnic groups within a racial category (e.g., Chinese, Korean, Japanese, and Vietnamese among Asians) often have different surnames and exhibit geographic clustering in their residential locations.  See \citet{greengard2025raking} for a raking method that partially relaxes this assumption.

Under this conditional independence assumption, the application of Bayes' rule yields the following posterior prediction of individual race \citep{elliott2009using}:
\begin{equation}
P(R = r \mid G = g, S = s) = \begin{cases} 
\frac{P(R = r \mid G = g) P(R = r \mid S = s)/P(R = r)}{\sum^K_{r'=1} P(R  = r' \mid G = g) P(R = r' \mid S = s)/P(R = r')} & \text{if } \ s \in \cS_{M}, \\
P(R = r \mid G = g) & \text{if } \ s \in \cS_{U}.
\end{cases}
\label{eq:bisg}
\end{equation}
where for uncommon surnames, substituting the marginal $P(R)$ for $P(R \mid S)$ in the BISG formula causes the surname term to cancel, leaving the prediction based entirely on the geographic racial composition $P(R \mid G)$. We next explain how to estimate an informative racial prior $P(R \mid S)$ for uncommon surnames that are not listed on the Census surname-by-race frequency table.

BIFSG extends BISG by incorporating first names \citep{voicu2018firstname}.  For the first time, the 2020 Census released the first name table, from which we can compute $P(R \mid F)$ for frequently occurring first names $\cF_M$.  We use $\cF_U$ to denote the set of first names that are not on the Census list.  Under the additional assumption of joint independence among surname, first name, and geolocation given race, we have the following BIFSG formula, 
\begin{equation}
\begin{aligned}
 & P(R = r \mid F = f, G = g, S = s) \\
= &  \begin{cases} 
\frac{P(R = r \mid G = g) P(R = r \mid S = s)/P(R = r)}{\sum^K_{r'=1} P(R = r' \mid G = g) P(R = r' \mid S = s)/ P(R = r')} & \text{if } \ s \in \cS_{M},\ f \in \cF_U, \\
P(R = r \mid G = g) & \text{if } \ s \in \cS_{U},\ f \in \cF_U, \\
\frac{P(R = r \mid G = g) P(R = r \mid F = f) P(R = r \mid S = s)/P(R = r)^2}{\sum^K_{r'=1} P(R = r' \mid G = g) P(R = r' \mid F = f) P(R = r' \mid S = s) / P(R = r')^2 } & \text{if } \ s \in \cS_M,\ f \in \cF_M, \\
\frac{P(R = r \mid G = g) P(R = r \mid F = f)/P(R = r)}{\sum^K_{r'=1} P(R = r' \mid G = g) P(R = r' \mid F = f)/P(R = r')} & \text{if } \ s \in \cS_{U},\ f \in \cF_M.
\end{cases}
\end{aligned}
\label{eq:bifsg}
\end{equation}

\subsection{Probabilistic Prediction based on Embeddings}

The proposed eBISG methodology proceeds in two steps. We first represent each name as a dense vector using a pre-trained multilingual text embedding model, which maps character strings into high-dimensional space. Because these models are trained on billions of text pairs and capture morphological patterns, names that share linguistic roots or etymological origins (including names that are not explicitly included in the training corpora) receive similar vector representations. We use the open-source E5-Large model \citep{wang2024e5}, which produces 1024-dimensional embeddings. We chose E5-Large because of its strong performance on cross-lingual retrieval benchmarks and its public availability, although the method generalizes to alternative embeddings models.  We verify in Appendix Section~\ref{sec:embedding_robustness} that results are qualitatively similar across alternative embedding models.

We then train feedforward neural network models on the Census name table to separately estimate $P(R \mid S)$ and $P(R \mid F)$ from name embeddings.  We use the fitted model to estimate race probabilities for names that do not appear in the Census data, and simply substitute these estimates into the BISG and BIFSG formulas in place of the uninformative marginal prior. We select the network architecture and training hyperparameters (number and width of hidden layers, dropout rate, learning rate, and batch size) using Optuna \citep{akiba2019optuna}, an automated Bayesian hyperparameter optimization framework, with 50 trials per model and an 80/20 train-validation split on the Census name table. The output is a six-class probability distribution over the Census racial categories.

Specifically, we train three versions of this eBISG model, which differ in their training data and the name representation that they receive as input. The surname embedding model is trained on the Census surname-by-race frequency table, using the 156,620 surnames and their associated race count distributions as training targets. The optimal architecture selected by Optuna for this model has two hidden layers of 1024 units each with dropout of 0.12. We then use the BISG formula in Equation~\eqref{eq:bisg}.
The surname and first name embedding version trains a separate model on the 53,616 Census first names and their race distributions, with Optuna selecting a two-layer architecture of 128 and 256 units.  We then use the BIFSG formula in Equation~\eqref{eq:bifsg}. The advantage of these two versions is that they require only publicly available Census data.

The full-name embedding version is trained on individual voter records with self-reported race from state voter registration files. We concatenate the first, middle, and last name of each voter into a single string before embedding, which allows the model, unlike BIFSG, to capture interactions between name components that are otherwise precluded by the conditional independence assumption. We train on a sample of two million voter records drawn from four Southern states (Alabama, Georgia, Louisiana, and South Carolina) and evaluate on held-out voter files from North Carolina and Florida. This ensures complete separation between training and evaluation data. Optuna selects a two-layer architecture of 1024 units each with dropout of 0.22 for this model.

\section{Empirical Validation}

We evaluate our eBISG methods using random samples of one million voters each from the North Carolina and Florida voter registration files, where self-reported race provides ground truth. The full-name embedding model is trained on two million voters drawn from four other Southern states (Alabama, Georgia, Louisiana, and South Carolina), ensuring complete separation between training and evaluation data. This also represents a hard test, requiring the generalization of results obtained in one population to another. We report results pooled across both states; state-by-state figures appear in Appendix Section~\ref{sec:bystate}. The results are qualitatively similar across the two states, with the ranking of methods preserved in every case.

\subsection{Race Prediction for Unmatched Voters}

\begin{figure}[t]
\centering
\includegraphics[width=\textwidth]{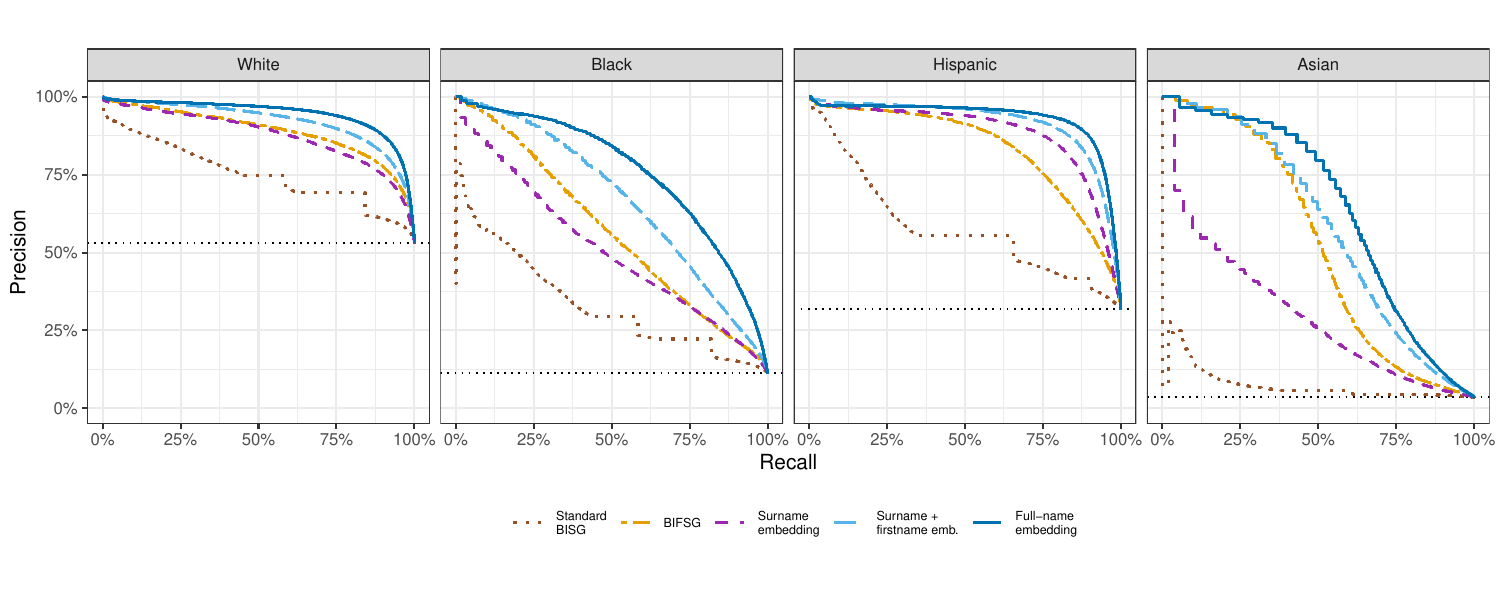}
\vspace{-.25in}
\caption{Precision-recall curves for BISG and eBISG race prediction among voters with unmatched surnames, pooling North Carolina and Florida. The dotted horizontal line indicates the baseline classifier (prevalence of each race among unmatched voters). Five methods are compared: standard BISG, BIFSG, surname embedding, surname and first name embedding, and full-name embedding.}
\label{fig:pr}
\end{figure}

Figure~\ref{fig:pr} presents precision-recall curves for voters with unmatched surnames under all five methods. Standard BISG (dotted red) performs worst, barely rising above the baseline classifier for Hispanic and Asian voters. Adding first name probabilities via BIFSG (yellow) produces a substantial improvement across all racial categories, confirming that first names carry independent racial information even when the surname is unlisted. 

The surname embedding (purple), which replaces the generic prior with an embedding-based prediction for unlisted surnames, achieves substantial gains over only surname information. For Hispanics, this model substantially outperforms BIFSG, indicating that the surname embeddings are quite informative for this group.
The surname and first name embedding (light blue) combines both embedding models and consistently improves upon BIFSG and the surname embedding approach, demonstrating that the embedding approach adds value beyond what the Census first name table provides even without access to voter file training data.
In particular, the first name embeddings bring a substantial gain in predictive power beyond the surname embeddings for Blacks and Asians.

The full-name embedding (dark blue) achieves the highest precision at every recall level for every race, by modeling the full name as a single representation that captures interactions between name components. Again, for Blacks and Asians, this gain is substantial.  State-by-state precision-recall results appear in Figure~\ref{fig:pr_bystate} of Appendix Section~\ref{sec:bystate}. Table~\ref{tab:examples} in Appendix Section~\ref{sec:additional} presents illustrative full-name embedding predictions for individual voters, and Table~\ref{tab:mae} reports tract-level mean absolute error by race and state.  These results are qualitatively similar to those of the pooled analysis presented here.

\subsection{Prediction Quality across the Income Distribution}

\begin{figure}[t]
\centering
\includegraphics[width=\textwidth]{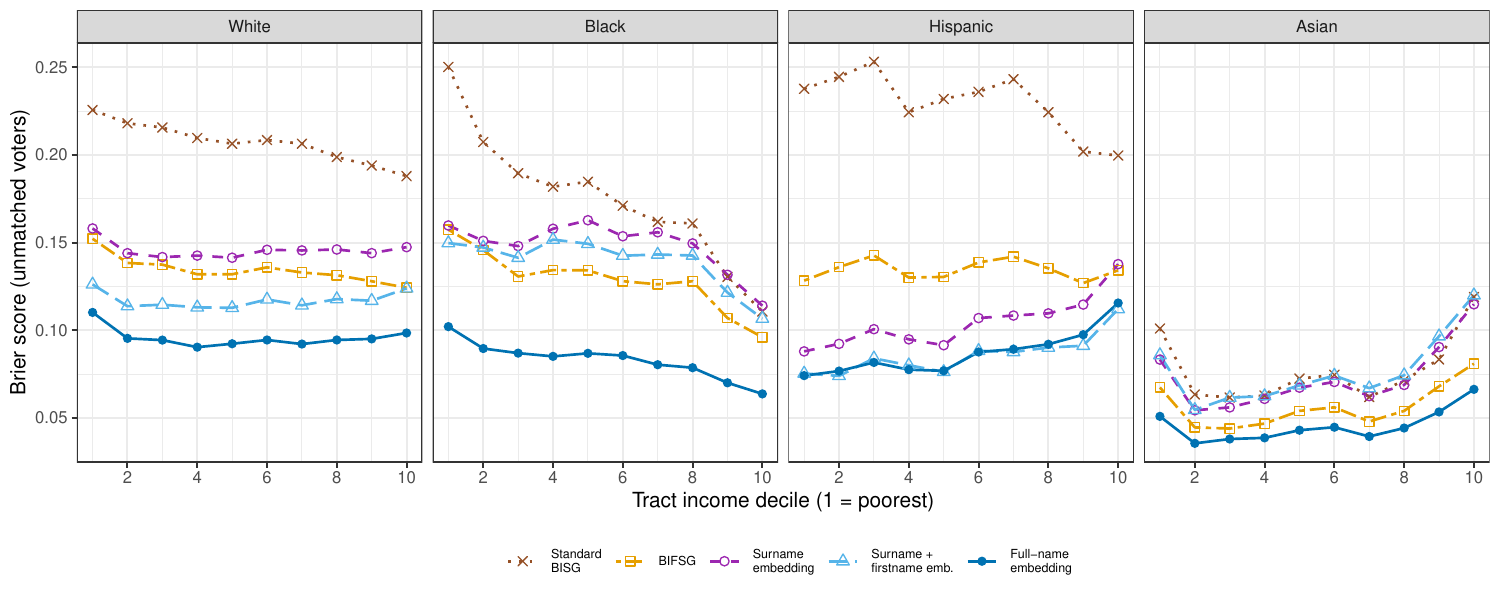}
\vspace{-.25in}
\caption{Brier score for individual-level race prediction among voters with unmatched surnames, by tract income decile. Tract-level Brier scores are averaged within each decile, weighted by the number of voters of the relevant race. Lower values indicate better prediction.}
\label{fig:brier_unmatched}
\end{figure}

A central concern raised by \citet{argylebarber2024} is that BISG misclassification is systematically correlated with neighborhood socioeconomic characteristics, biasing downstream disparity estimates. Figure~\ref{fig:brier_unmatched} examines this question directly by plotting Brier scores for unmatched voters against tract-level median household income (see Figure~\ref{fig:brier_bystate} for state-by-state results). Within each tract, we compute the Brier score as the mean squared difference between the predicted race probability and the binary race indicator across all unmatched voters. We then take a weighted average across tracts within each income decile, weighting by the number of voters of the relevant race in each tract.

Standard BISG exhibits a pronounced income gradient. Brier scores are highest in the poorest tracts for White voters and in the wealthiest tracts for Black and Hispanic voters, consistent with the pattern documented by \cite{argylebarber2024}. All four alternatives to standard BISG reduce Brier scores, with the full-name embedding achieving the lowest scores across all income deciles. The systematic, income-correlated misclassification that concerns \cite{argylebarber2024} appears to be at least in large part a consequence of the uninformative generic prior rather than a fundamental limitation of the BISG framework. Once the prior is replaced with an informative name-based prediction, the income gradient substantially attenuates.

\subsection{Calibration}

\begin{figure}[!t]
\centering
\includegraphics[width=\textwidth]{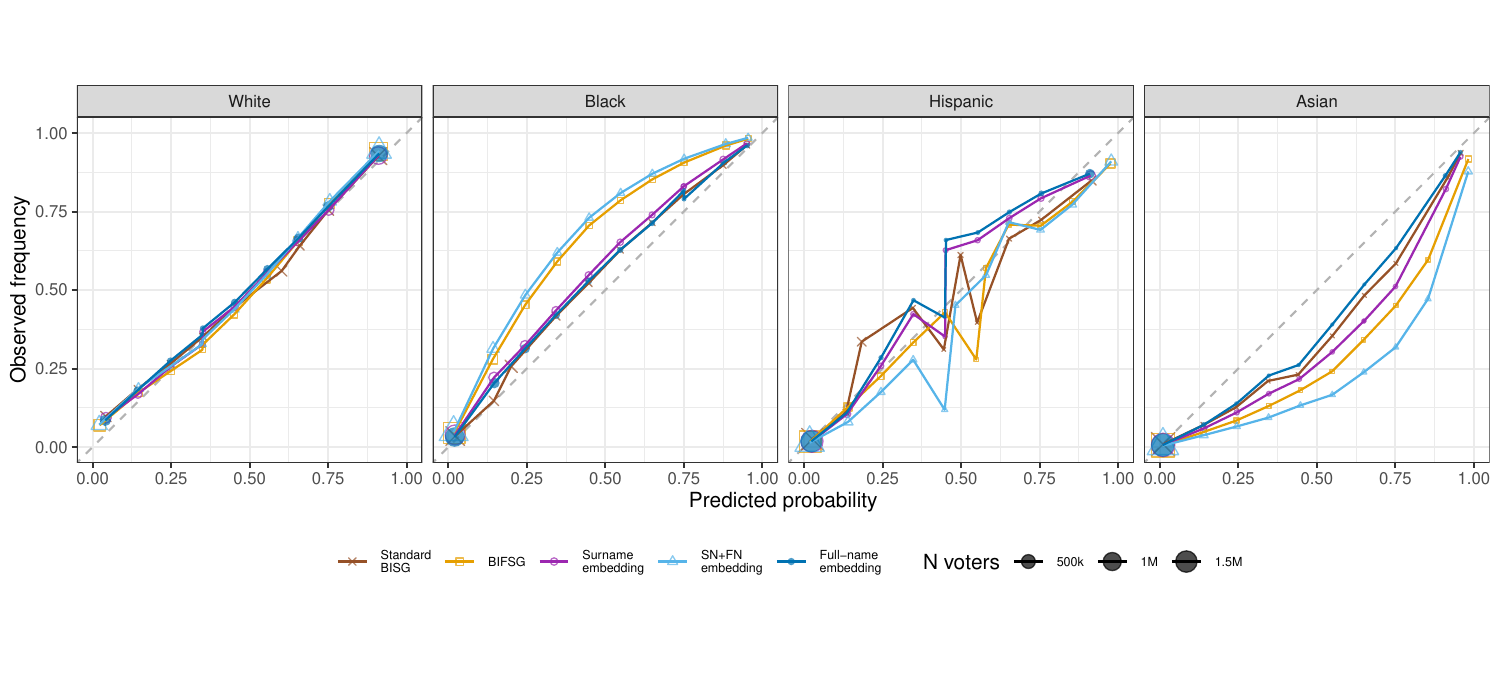}
\vspace{-.5in}
\caption{Calibration of BISG race probabilities for all voters. Points are sized by the number of voters in each predicted-probability bin. A perfectly calibrated model places all points on the 45-degree line.}
\label{fig:calibration}
\end{figure}

Beyond discrimination, the calibration of predicted probabilities matters for downstream analyses that use BISG posteriors as inputs \citep{mccartan2025birdie}. Figure~\ref{fig:calibration} plots predicted probabilities against observed frequencies for all voters across all five methods. A perfectly calibrated model would place all points on the 45-degree line. All methods are reasonably well-calibrated, but the full-name embedding model provides the best calibration across all racial groups. The results imply that the full-name embedding model is able to achieve the highest precision and recall of racial prediction while also improving calibration.  For completeness, we also provide the calibration results among unmatched voters (see Figure~\ref{fig:calibration_unmatched} of Appendix Section~\ref{sec:additional}), where the full-name embedding achieves the best calibration for Asian voters while BIFSG is best calibrated for Black voters. However, these results among unmatched voters are of less interest since calibration is an aggregate-level property and most researchers would not analyze unmatched voters in isolation.

\subsection{Full-Name Embedding for Matched Voters}

The results above apply the eBISG methods only to voters with unmatched surnames, preserving the Census-derived $P(R \mid S)$, and $P(R \mid F)$ whenever appropriate, for matched voters. However, the full-name embedding can also be applied to all voters, replacing the Census surname and first name priors with a prediction that incorporates the entire information based on first, middle, and last names. Standard BISG assigns identical priors to two voters with the same last name but racially distinctive first names despite the additional signal in their given names. BIFSG takes into account first names, but it does so only for frequently occurring first names.  Moreover, BIFSG ignores middle names and does not consider the interaction between first and last names. The full-name embedding captures all such signal and has the potential to improve predictions even for voters whose surnames are already in the Census surname-by-race frequency table.

Figure~\ref{fig:overall_matched} of Appendix Section~\ref{subsec:overall_results} evaluate three methods for matched voters, pooling North Carolina and Florida: standard BISG, BIFSG, and full-name embedding. We omit the surname embedding and surname-plus-first name embedding because both produce identical predictions to standard BISG for matched voters by construction. While BIFSG improves by incorporating first name information, the full-name embedding model improves further across most racial categories and income deciles. As expected, the improvement is modest relative to the gains for unmatched voters. The Census surname-by-race frequency table is already sufficiently informative for matched names, and the additional signal from given names is incremental. The full-name model is trained on voter files from four Southern states whose racial composition differs from the evaluation states, and this distributional shift limits the performance gains, in particular for Hispanic voters in Florida. Researchers with access to voter file data that better matches their target population could expect larger improvements. State-by-state results for matched voters appear in Figures~\ref{fig:pr_matched_bystate} and~\ref{fig:brier_matched_bystate} of Appendix Section~\ref{subsec:overall_results}.

\section{Concluding Remarks}

BISG is the workhorse approach for racial imputation in the social sciences. However, roughly one in ten Americans hold a surname that is not found in the Census surname-by-race frequency table. For these individuals, BISG falls back on an uninformative prior that carries no information about the name at all. This disproportionately affects Hispanic and Asian voters, whose surnames are relatively more rare and thus excluded from the table, and suggests that the method is weakest precisely where accurate measurement of minority populations is most important.  

Pre-trained text embedding models offer one option for filling this gap without departing from the assumptions and approach of the BISG framework. By representing names as dense vectors that encode morphological and linguistic structure, surnames that do not appear in the Census surname-by-race frequency table can nonetheless receive informative race probabilities based on their semantic similarities to names that are represented. When the 2020 Census first name data are also incorporated, the performance gains extend further. A full-name embedding model trained on voter file data improves performance even further, by capturing interactions between given names and surnames that cannot be represented by the conditional independence assumption underlying BIFSG. Because our embedding-based methods leave Census-matched predictions unchanged, researchers adopting the approach do not face the risk of degrading the quality of predictions already handled well by BISG.  

The proposed eBISG approach does have limitations that warrant emphasis. We trained our full-name embedding model on voter files from four Southern states whose racial compositions differ from the two evaluation states. This distributional shift limits generalization, which we observe for Hispanic voters in Florida whose prevalence is higher than in the training population. Researchers seeking to apply the full-name model outside of the South should expect some attenuation of gains relative to the performance that we report. The surname embedding and surname-plus-first name embedding methods, which rely only on Census data, do not face similar constraints. 

The conditional independence assumption $F \indep S \mid R$, which determines how first name and surname information are combined in BIFSG and in our surname-plus-first name embedding, is likely violated in practice. This is particularly the case within the Asian category, where distinct ethnic groups carry distinctive name pairings. The advantage of the full-name embedding in this setting reflects in part its ability to capture these violations. In addition to raking \citep{greengard2025raking}, future work could explore fine-tuning the embedding model on local voter file data where available, or incorporating additional name information such as suffixes into the embedding representation.

Finally, the eBISG method that we propose is designed to integrate with the existing ecosystem of tools built on BISG posteriors, including fBISG \citep{imai2022fbisg} and BIRDiE \citep{mccartan2025birdie}, without requiring modifications to those downstream approaches. Towards this end, we provide open-source implementations of all models as well as the trained weights, allowing applied researchers to adopt the approach by replacing a single input to their existing BISG pipeline.

\bigskip
\bibliography{references}

@Manual{wru,
  title = {wru: Who are You? Bayesian Prediction of Racial Category Using Surname, First Name, Middle Name, and Geolocation},
  author = {Kabir Khanna and Brandon Bertelsen and Santiago Olivella and Evan Rosenman and Alexander Rossell Hayes and Kosuke Imai},
  year = {2024},
  note = {R package version 3.0.3},
  url = {https://CRAN.R-project.org/package=wru},
  doi = {10.32614/CRAN.package.wru}
}

@article{greenwald2024regulatory,
  title={Regulatory Arbitrage or Random Errors? {I}mplications of Race Prediction Algorithms in Fair Lending Analysis},
  author={Greenwald, Daniel and Howell, Sabrina T. and Li, Cangyuan and Yimfor, Emmanuel},
  journal={Journal of Financial Economics},
  volume={157},
  pages={103857},
  year={2024}
}

@book{fraga2018turnout,
  title={The Turnout Gap: Race, Ethnicity, and Political Inequality in a Diversifying America},
  author={Fraga, Bernard L.},
  year={2018},
  publisher={Cambridge University Press}
}

@article{adjayegbewonyo2014bisg,
  title={Using the {B}ayesian Improved Surname Geocoding Method ({BISG}) to Create a Working Classification of Race and Ethnicity in a Diverse Managed Care Population: A Validation Study},
  author={Adjaye-Gbewonyo, Dzifa and Bednarczyk, Robert A. and Davis, Rachel L. and Omer, Saad B.},
  journal={Health Services Research},
  volume={49},
  number={1},
  pages={268--283},
  year={2014}
}

@article{fiscella2006geocoding,
  title={Use of Geocoding and Surname Analysis to Estimate Race and Ethnicity},
  author={Fiscella, Kevin and Fremont, Allen M.},
  journal={Health Services Research},
  volume={41},
  number={4p1},
  pages={1482--1500},
  year={2006}
}

@article{elliott2009using,
  title={Using the {C}ensus {B}ureau's Surname List to Improve Estimates of Race/Ethnicity and Associated Disparities},
  author={Elliott, Marc N. and Morrison, Peter A. and Fremont, Allen and McCaffrey, Daniel F. and Pantoja, Philip and Lurie, Nicole},
  journal={Health Services and Outcomes Research Methodology},
  volume={9},
  number={2},
  pages={69--83},
  year={2009}
}

@techreport{cfpb2014proxy,
  title={Using Publicly Available Information to Proxy for Unidentified Race and Ethnicity},
  author={{Consumer Financial Protection Bureau}},
  year={2014},
  institution={CFPB},
  address={Washington, DC}
}

@article{imaikhanna2016,
  title={Improving Ecological Inference by Predicting Individual Ethnicity from Voter Registration Records},
  author={Imai, Kosuke and Khanna, Kabir},
  journal={Political Analysis},
  volume={24},
  number={2},
  pages={263--272},
  year={2016}
}

@article{imai2022fbisg,
  title={Addressing Census Data Problems in Race Imputation via Fully {B}ayesian Improved Surname Geocoding and Name Supplements},
  author={Imai, Kosuke and Olivella, Santiago and Rosenman, Evan T. R.},
  journal={Science Advances},
  volume={8},
  number={49},
  pages={eadc9824},
  year={2022}
}

@article{deluca2024power,
  title={The Power of Characters: Evaluating Machine Learning-Modified {B}ayesian Improved Surname Geocoding Inference of Race in Redistricting},
  author={DeLuca, Kevin and Curiel, John A.},
  journal={State Politics \& Policy Quarterly},
  volume={24},
  number={3},
  pages={300--321},
  year={2024}
}

@article{wang2024e5,
  title={Multilingual {E5} Text Embeddings: A Technical Report},
  author={Wang, Liang and Yang, Nan and Huang, Xiaolong and Yang, Linjun and Majumder, Rangan and Wei, Furu},
  journal={arXiv preprint arXiv:2402.05672},
  year={2024}
}

@article{mccartan2025birdie,
  title={Estimating Racial Disparities When Race Is Not Observed},
  author={McCartan, Cory and Fisher, Robin and Goldin, Jacob and Ho, Daniel E. and Imai, Kosuke},
  journal={Journal of the American Statistical Association},
  volume={120},
  number={552},
  pages={2140--2153},
  year={2025}
}

@article{argylebarber2024,
  title={Misclassification and Bias in Predictions of Individual Ethnicity from Administrative Records},
  author={Argyle, Lisa P. and Barber, Michael J.},
  journal={American Political Science Review},
  volume={118},
  number={2},
  pages={1058--1066},
  year={2024}
}

@article{sood2018predicting,
  title={Predicting Race and Ethnicity from the Sequence of Characters in a Name},
  author={Sood, Gaurav and Laohaprapanon, Suriyan},
  journal={arXiv preprint arXiv:1805.02109},
  year={2018}
}

@article{parasurama2021racebert,
  title={race{BERT}: A Transformer-Based Model for Predicting Race and Ethnicity from Names},
  author={Parasurama, Prasanna},
  journal={arXiv preprint arXiv:2112.03807},
  year={2021}
}

@article{jain2022importance,
  title={The Importance of Being {E}rnest, {E}kundayo, or {E}swari: An Interpretable Machine Learning Approach to Name-Based Ethnicity Classification},
  author={Jain, Vaishali and Enamorado, Ted and Rudin, Cynthia},
  journal={Harvard Data Science Review},
  volume={4},
  number={3},
  year={2022}
}

@inproceedings{ye2017nationality,
  title={Nationality Classification Using Name Embeddings},
  author={Ye, Junting and Han, Shuchu and Hu, Yifan and Coskun, Baris and Liu, Meizhu and Qin, Hong and Skiena, Steven},
  booktitle={Proceedings of the 2017 ACM Conference on Information and Knowledge Management},
  pages={1897--1906},
  year={2017}
}

@inproceedings{ye2019secret,
  title={The Secret Lives of Names? {N}ame Embeddings from Social Media},
  author={Ye, Junting and Skiena, Steven},
  booktitle={Proceedings of the 25th ACM SIGKDD International Conference on Knowledge Discovery \& Data Mining},
  pages={1--9},
  year={2019}
}

@article{voicu2018firstname,
  title={Using First Name Information to Improve Race and Ethnicity Classification},
  author={Voicu, Ioan},
  journal={Statistics and Public Policy},
  volume={5},
  number={1},
  pages={1--13},
  year={2018}
}

@article{rosenman2023names,
  title={Race and Ethnicity Data for First, Middle, and Surnames},
  author={Rosenman, Evan T. R. and Olivella, Santiago and Imai, Kosuke},
  journal={Scientific Data},
  volume={10},
  pages={299},
  year={2023}
}

@article{leevelez2025,
  title={Measuring Descriptive Representation at Scale: Methods for Predicting the Race and Ethnicity of Public Officials},
  author={Lee, Diana Da In and Velez, Yamil Ricardo},
  journal={British Journal of Political Science},
  volume={55},
  pages={e110},
  year={2025}
}

@article{greengard2025raking,
  title={A Calibrated {BISG} for Inferring Race from Surname and Geolocation},
  author={Greengard, Philip and Gelman, Andrew},
  journal={Journal of the Royal Statistical Society Series A: Statistics in Society},
  year={2025}
}

@inproceedings{akiba2019optuna,
  title={Optuna: A Next-Generation Hyperparameter Optimization Framework},
  author={Akiba, Takuya and Sano, Shotaro and Yanase, Toshihiko and Ohta, Takeru and Koyama, Masanori},
  booktitle={Proceedings of the 25th ACM SIGKDD International Conference on Knowledge Discovery \& Data Mining},
  pages={2623--2631},
  year={2019}
}

\bigskip
\appendix
\begin{center}
{\bf \LARGE Supplementary Appendix}
\end{center}
\renewcommand{\thetable}{A\arabic{table}}
\renewcommand{\thefigure}{A\arabic{figure}}
\setcounter{table}{0}
\setcounter{figure}{0}

\FloatBarrier
\section{State-by-State Results}
\label{sec:bystate}

\begin{table}[!h]
\centering
\small
\begin{tabular}{l l rr rr}
\toprule
 & & \multicolumn{2}{c}{Surname matched} & \multicolumn{2}{c}{Surname unmatched} \\
\cmidrule(lr){3-4} \cmidrule(lr){5-6}
State & Race & FN matched & FN unmatched & FN matched & FN unmatched \\
\midrule
NC & White    & 71.5 & 0.9 & 60.0 & 1.4 \\
NC & Black    & 19.6 & 2.7 & 12.3 & 3.0 \\
NC & Hispanic &  3.4 & 0.2 & 20.3 & 1.4 \\
NC & Asian    &  1.5 & 0.4 &  3.3 & 1.4 \\
NC & AIAN     &  0.8 & 0.1 &  0.5 & 0.1 \\
NC & Other    &  1.5 & 0.2 &  4.2 & 0.9 \\
\midrule
FL & White    & 62.1 & 1.0 & 46.8 & 1.5 \\
FL & Black    & 12.5 & 2.4 &  7.1 & 2.2 \\
FL & Hispanic & 16.2 & 1.0 & 32.7 & 3.5 \\
FL & Asian    &  2.0 & 0.4 &  2.1 & 0.9 \\
FL & AIAN     &  0.3 & 0.0 &  0.2 & 0.0 \\
FL & Other    &  1.8 & 0.2 &  2.5 & 0.5 \\
\bottomrule
\end{tabular}
\caption{Racial composition of voters by surname and first name match status, by state. Percentages sum to 100 within each surname group.}
\label{tab:unmatched_bystate}
\end{table}

\begin{figure}[!h]
\centering
\includegraphics[width=\textwidth]{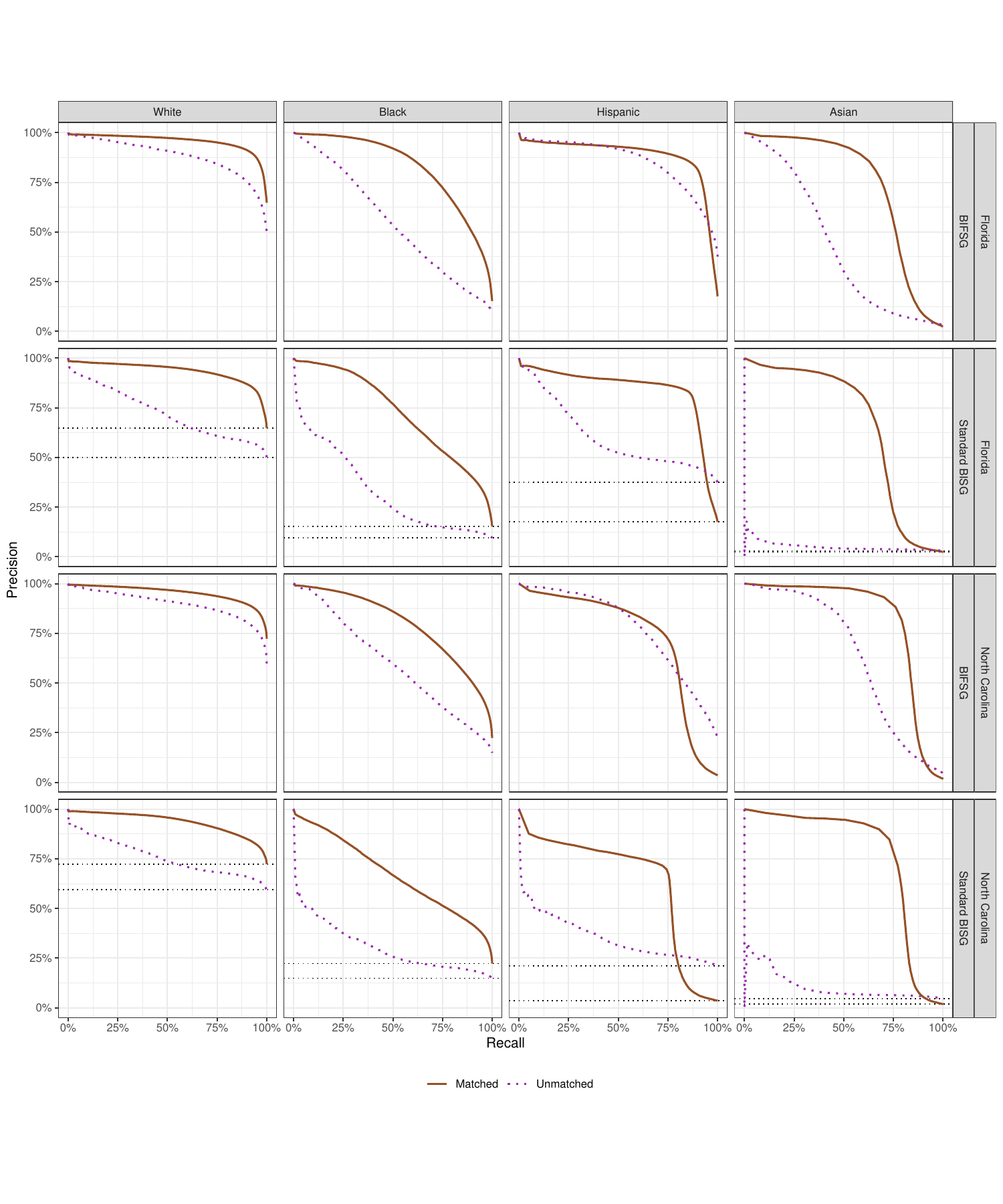}
\vspace{-.5in}
\caption{Precision-recall curves for standard BISG and BIFSG, matched versus unmatched voters, by state.}
\label{fig:pr_mv_bystate}
\end{figure}

\begin{figure}[!h]
\centering
\includegraphics[width=\textwidth]{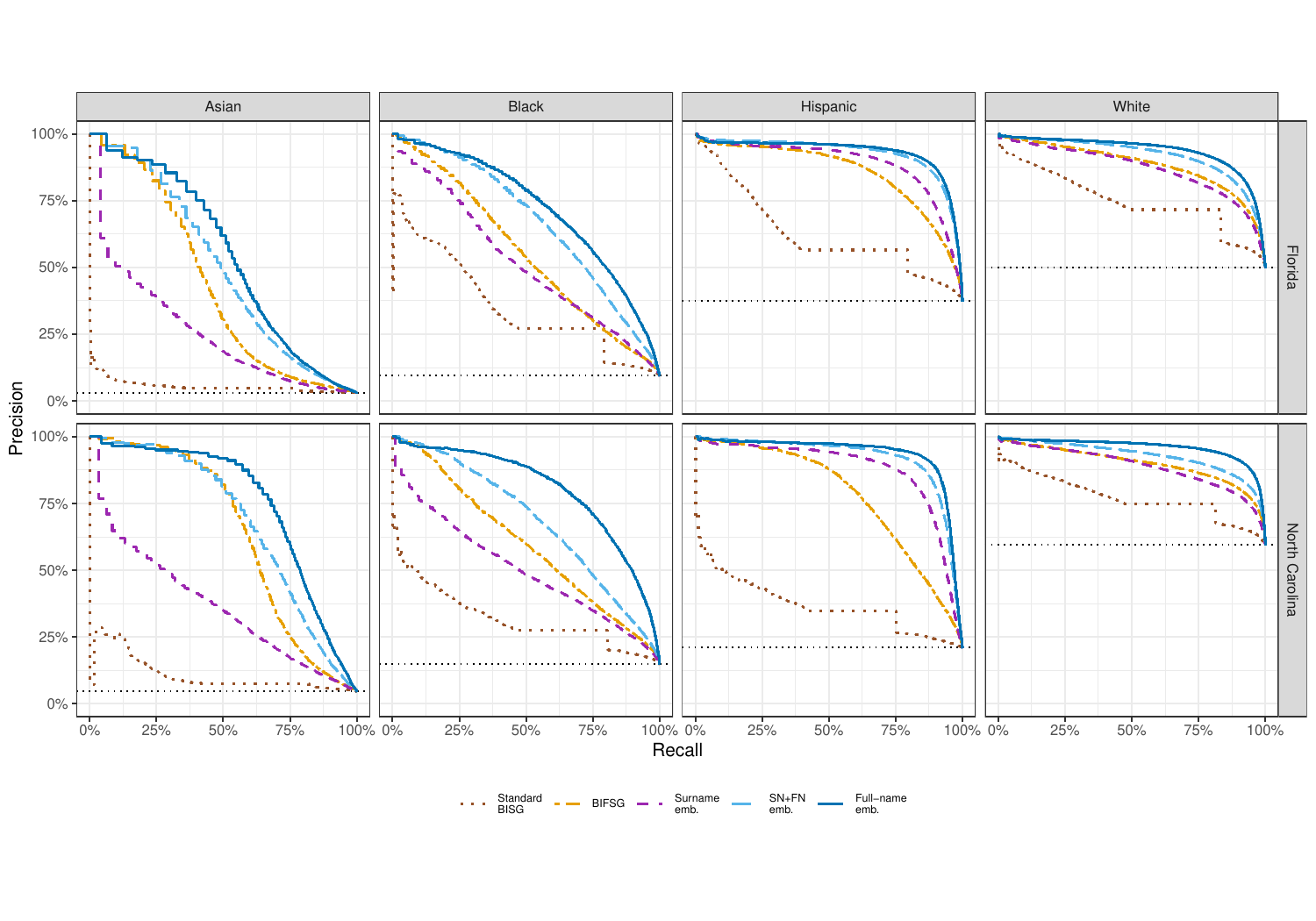}
\vspace{-.5in}
\caption{Precision-recall curves for unmatched voters, five methods, by state.}
\label{fig:pr_bystate}
\end{figure}

\begin{figure}[!h]
\centering
\includegraphics[width=\textwidth]{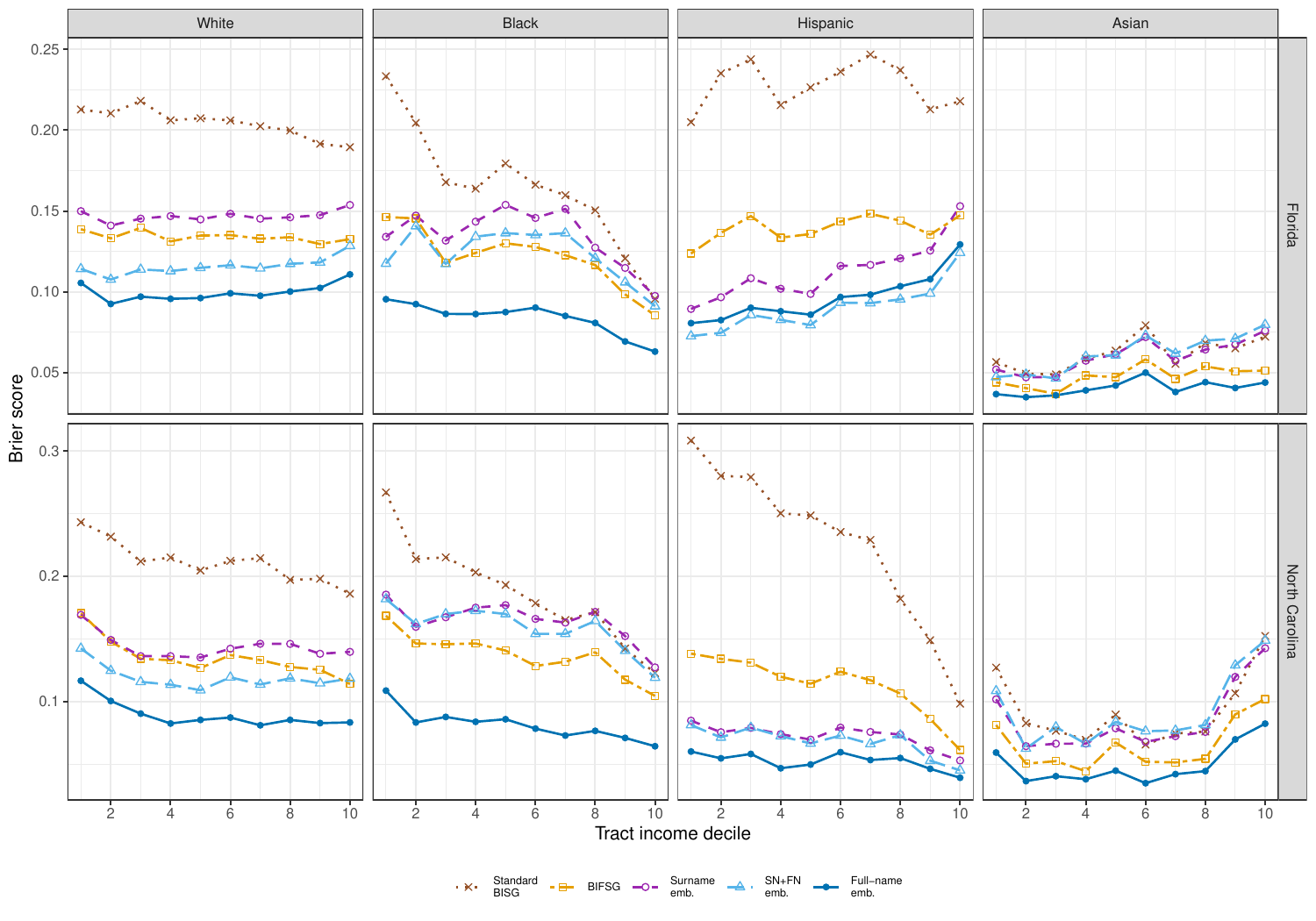}
\vspace{-.25in}
\caption{Brier score by income decile for unmatched voters, five methods, by state.}
\label{fig:brier_bystate}
\end{figure}

\FloatBarrier
\section{Matched Voters}

\subsection{Overall results}
\label{subsec:overall_results}

\begin{figure}[!h]
\begin{subfigure}[a]{\textwidth}
\centering
\includegraphics[width=\textwidth]{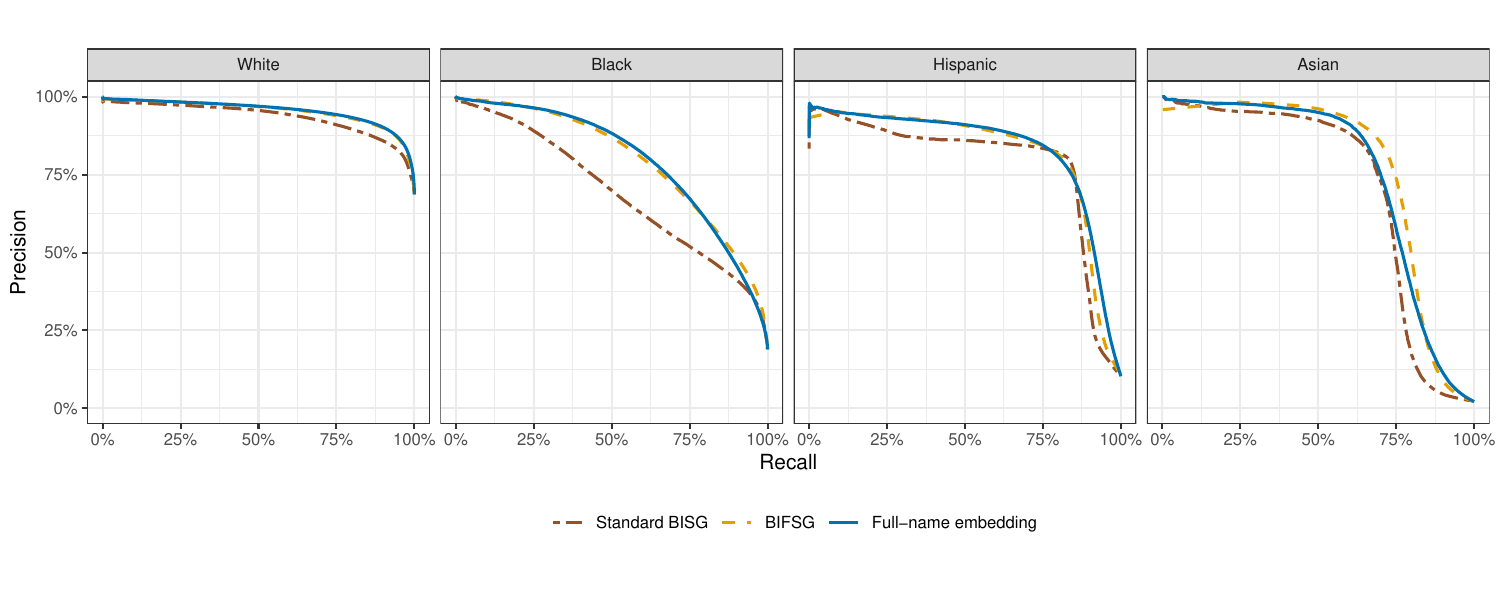}
\vspace{-.5in}
\caption{Precision-recall curves.}
\label{fig:pr_matched}
\end{subfigure}
\begin{subfigure}[c]{\textwidth}
\centering
\includegraphics[width=\textwidth]{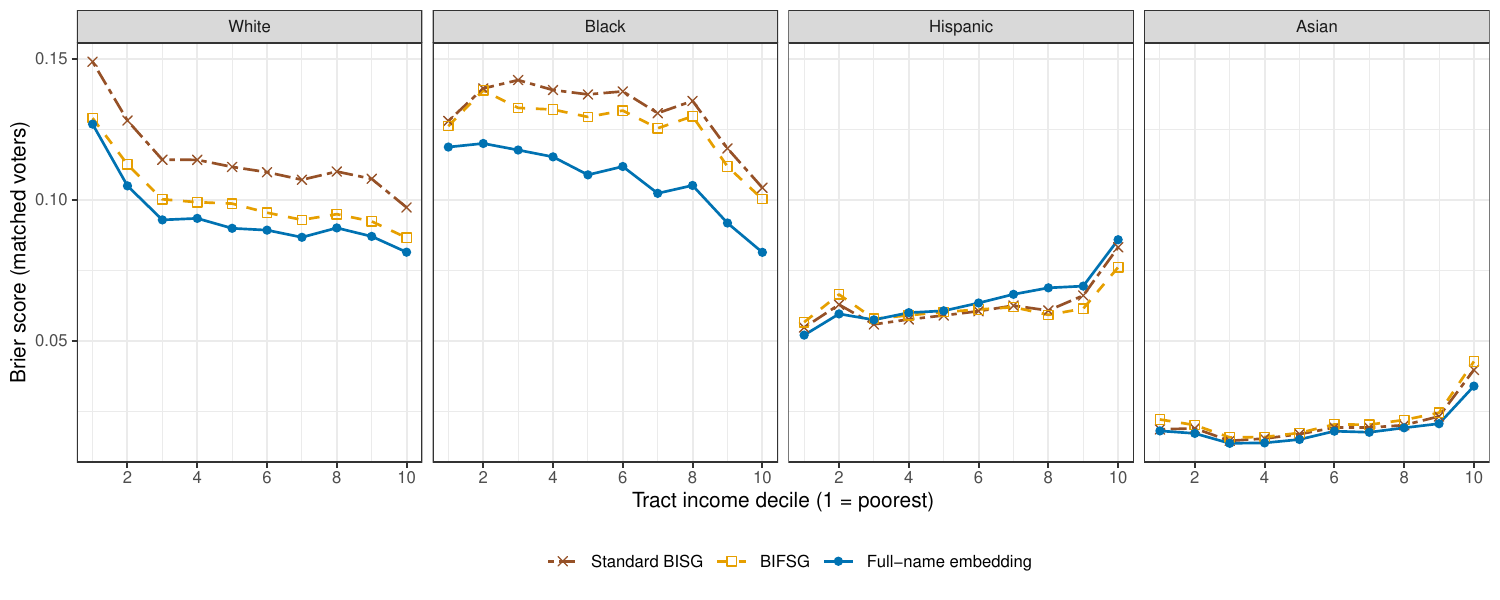}
\vspace{-.25in}
\caption{Brier score for matched voters.}
\label{fig:brier_matched}
\end{subfigure}
\caption{Predictive Performance of standard BISG, BIFSG, and full-name embedding approach for matched voters.} \label{fig:overall_matched}
\end{figure}

\clearpage
\FloatBarrier
\subsection{State by state results}

\begin{figure}[!h]
\centering
\includegraphics[width=\textwidth]{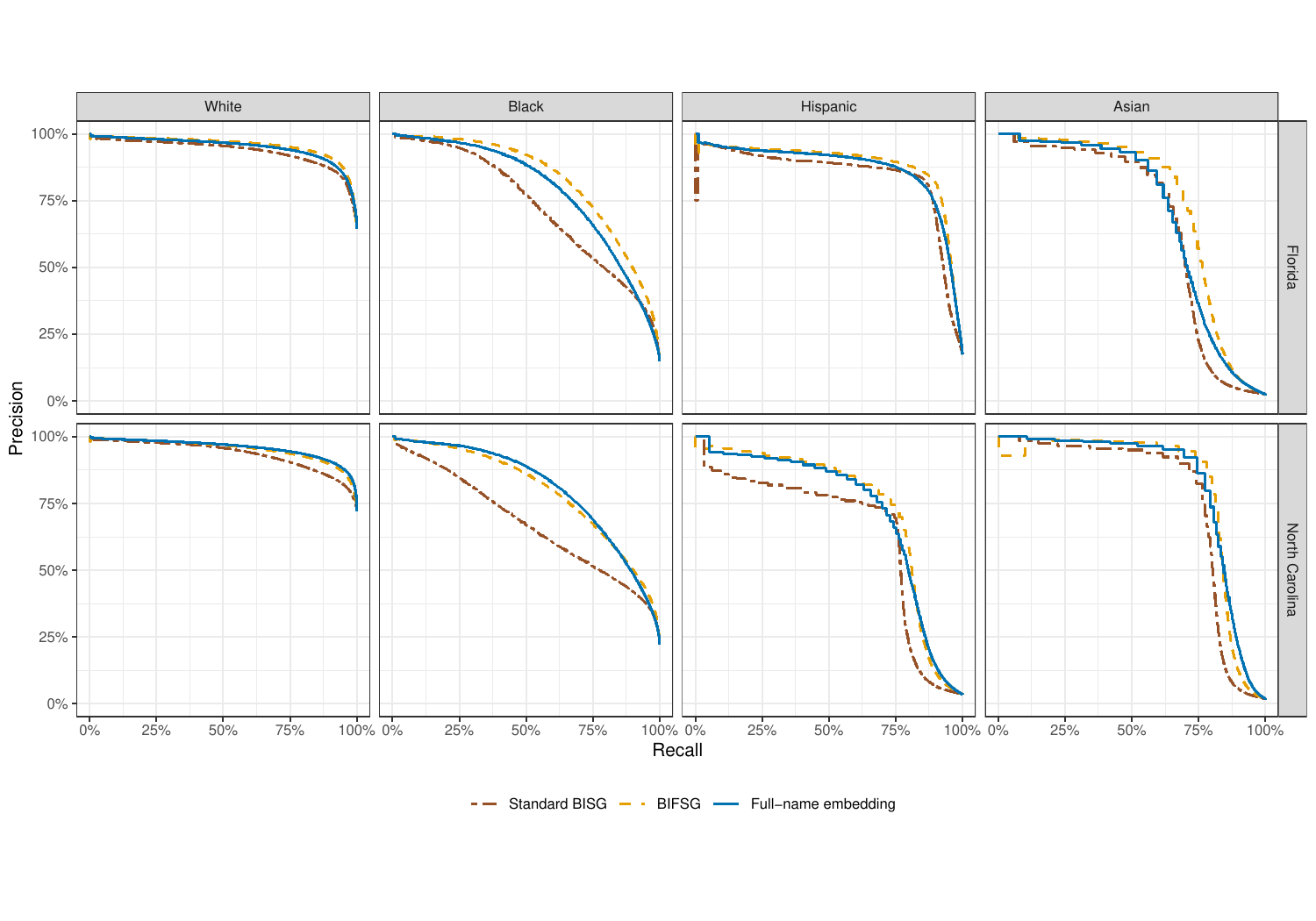}
\vspace{-.25in}
\caption{Precision-recall curves for matched voters, three methods, by state.}
\label{fig:pr_matched_bystate}
\end{figure}

\begin{figure}[!h]
\centering
\includegraphics[width=\textwidth]{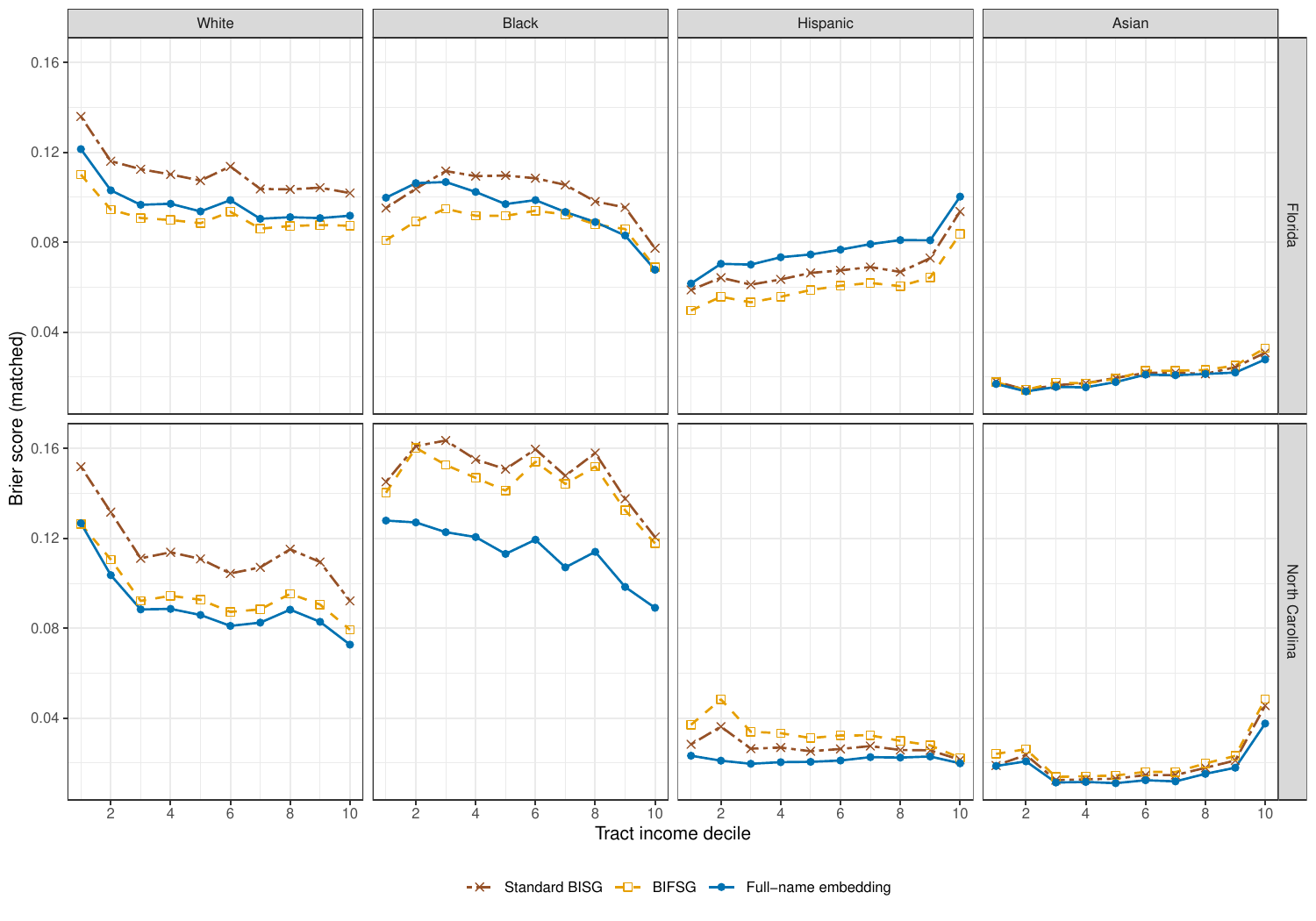}
\vspace{-.25in}
\caption{Brier score by income decile for matched voters, three methods, by state.}
\label{fig:brier_matched_bystate}
\end{figure}

\FloatBarrier
\section{Additional Results}
\label{sec:additional}

\begin{figure}[!h]
\centering
\includegraphics[width=\textwidth]{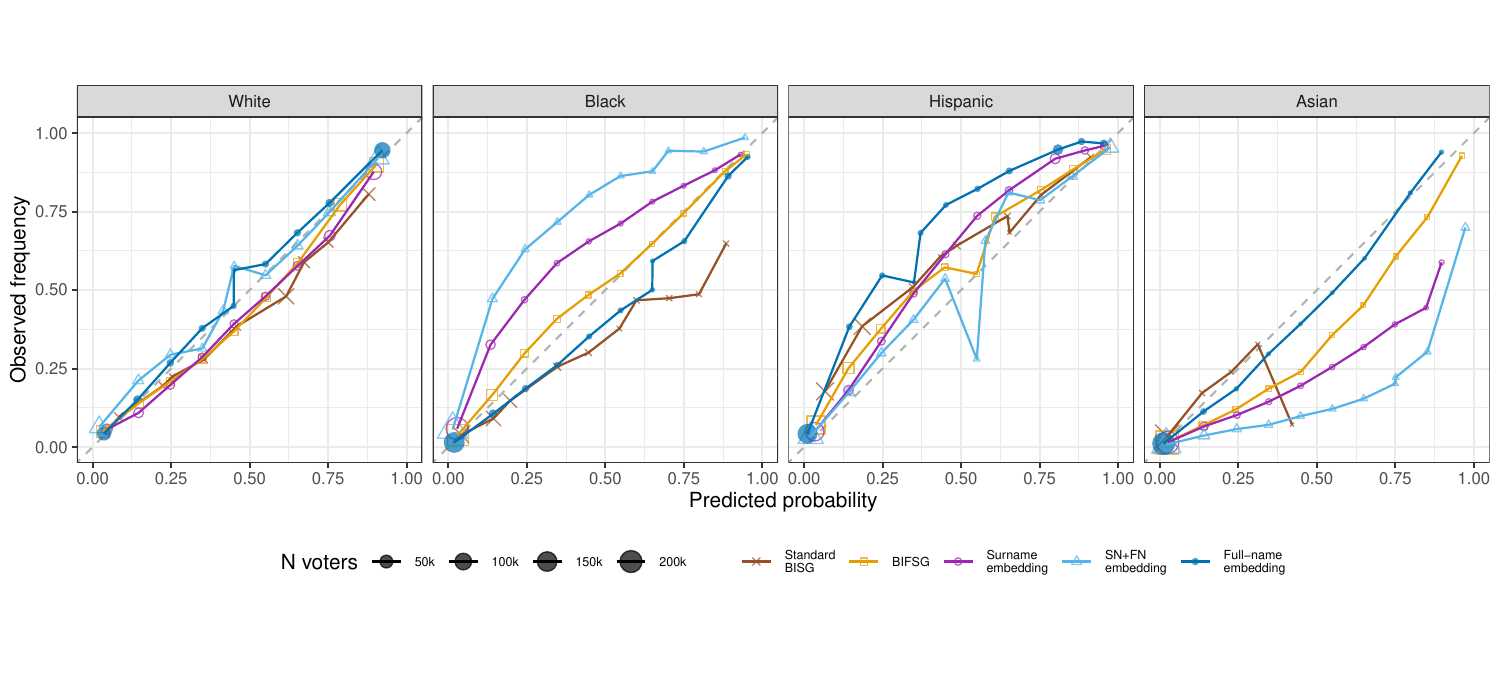}
\vspace{-.5in}
\caption{Calibration of BISG race probabilities for unmatched voters only. Among the embedding models, the full-name embedding achieves the best calibration for Asian voters, while BIFSG is best calibrated for Black voters.}
\label{fig:calibration_unmatched}
\end{figure}

\begin{table}[!h]
\centering
\small
\begin{tabular}{ll rrr rrr}
\toprule
 & & \multicolumn{3}{c}{All voters} & \multicolumn{3}{c}{Unmatched only} \\
\cmidrule(lr){3-5} \cmidrule(lr){6-8}
State & Race & Standard & Surname & Full-name & Standard & Surname & Full-name \\
\midrule
NC & White    & 8.21 & 8.09 & 7.87 & 13.69 & 11.82 & 7.64 \\
NC & Black    & 7.93 & 7.90 & 7.54 &  9.98 &  9.59 & 5.32 \\
NC & Hispanic & 4.54 & 4.36 & 3.96 & 12.24 &  9.58 & 3.03 \\
NC & Asian    & 1.23 & 1.19 & 1.03 &  6.15 &  5.46 & 2.75 \\
NC & AIAN     & 0.68 & 0.67 & 0.68 &  0.77 &  0.69 & 0.77 \\
NC & Other    & 0.95 & 0.95 & 0.94 &  3.77 &  3.80 & 3.41 \\
\midrule
FL & White    & 6.93 & 6.74 & 6.39 & 12.20 & 11.29 & 7.49 \\
FL & Black    & 4.93 & 4.86 & 4.72 &  6.08 &  5.43 & 5.12 \\
FL & Hispanic & 3.29 & 3.12 & 2.89 &  8.12 &  7.28 & 5.34 \\
FL & Asian    & 1.50 & 1.40 & 1.21 &  5.25 &  4.21 & 2.73 \\
FL & AIAN     & 0.48 & 0.44 & 0.48 &  0.82 &  0.41 & 0.86 \\
FL & Other    & 1.22 & 1.23 & 1.11 &  3.28 &  3.29 & 3.23 \\
\bottomrule
\end{tabular}
\caption{Mean absolute error (MAE) in tract-level racial composition estimates, in percentage points. For each tract with at least 20 voters in the relevant group, we compute the absolute difference between the mean BISG posterior and the true racial proportion.}
\label{tab:mae}
\end{table}

\begin{table}[!h]
\centering
\small
\begin{tabular}{ll l rrrr}
\toprule
First name & Last name & True race & $\hat{P}$(Whi) & $\hat{P}$(Bla) & $\hat{P}$(His) & $\hat{P}$(Asi) \\
\midrule
Susan     & Recenello       & White    & .989 & .002 & .002 & .001 \\
Susan     & Schainblatt     & White    & .988 & .001 & .001 & .001 \\
Susan     & Hechtlinger     & White    & .987 & .001 & .000 & .000 \\
Davion    & Kennedy-Wall    & Black    & .000 & .978 & .000 & .000 \\
Ky'Zion   & Collins-Andrews & Black    & .000 & .974 & .000 & .000 \\
Dyquavius & Johnson-Cates   & Black    & .000 & .974 & .000 & .000 \\
Ysaias    & Martinez-Vega   & Hispanic & .012 & .002 & .930 & .004 \\
Yessenia  & Reyes-Vidales   & Hispanic & .015 & .002 & .927 & .004 \\
Javier    & Jaimes-Ayala    & Hispanic & .012 & .003 & .926 & .003 \\
Chien     & H               & Asian    & .002 & .000 & .000 & .919 \\
Tom       & Va              & Asian    & .000 & .000 & .000 & .916 \\
Rin       & Y               & Asian    & .005 & .001 & .001 & .910 \\
\bottomrule
\end{tabular}
\caption{Illustrative full-name embedding predictions for North Carolina voters with surnames absent from the Census surname-by-race frequency table. Predicted probabilities are from the neural network prior to BISG geographic adjustment.}
\label{tab:examples}
\end{table}

\clearpage
\FloatBarrier
\section{Sensitivity to Embedding Model Choice}
\label{sec:embedding_robustness}

\begin{figure}[!h]
\centering
\includegraphics[width=\textwidth]{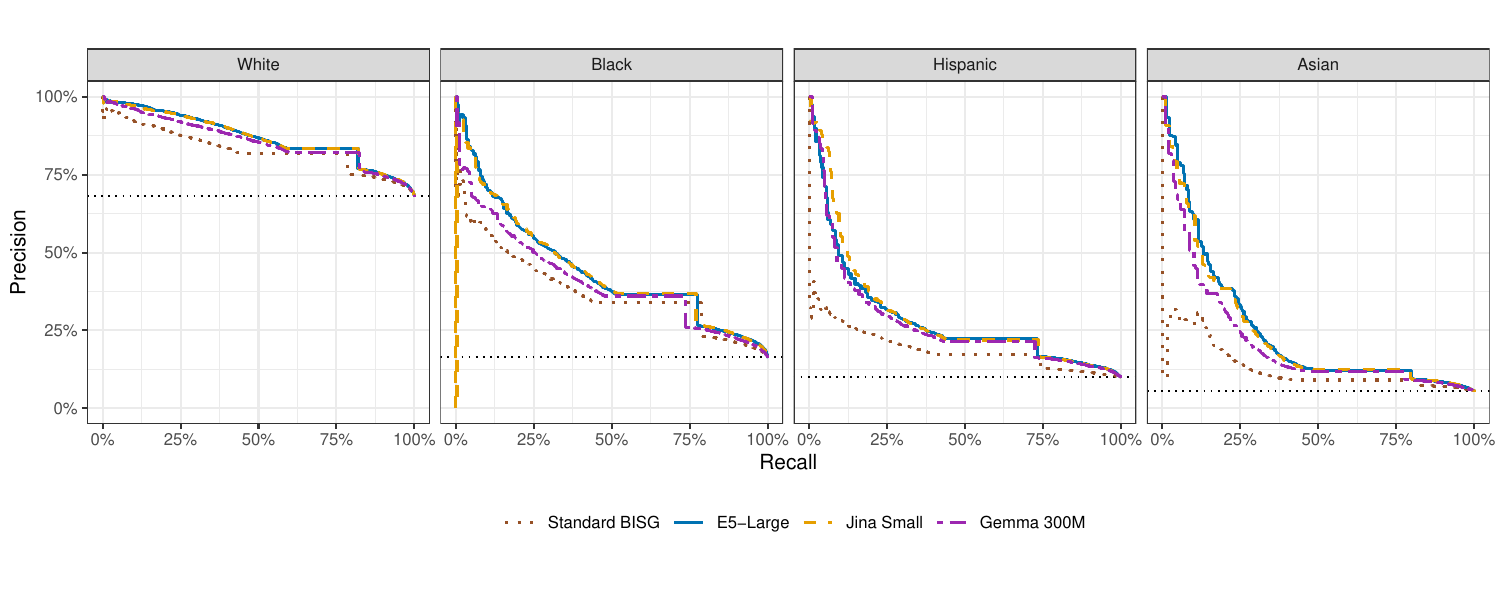}
\vspace{-.5in}
\caption{Precision-recall curves for unmatched voters in North Carolina under alternative pre-trained embedding models. All embedding models improve over standard BISG, with E5-Large achieving the best performance.}
\label{fig:embedding_comparison}
\end{figure}

The results reported in the main text use E5-Large \citep{wang2024e5} as the pre-trained embedding model. To assess sensitivity to this choice, we trained the surname embedding model using two alternative embedding architectures on the Census surname-by-race frequency table: Jina Small (512 dimensions, 33M parameters) and Gemma 300M (768 dimensions). Figure~\ref{fig:embedding_comparison} presents precision-recall curves for unmatched voters in North Carolina under each model alongside standard BISG. All embedding models produce substantial improvements over standard BISG, with E5-Large yielding the best performance across all racial categories. The ranking is consistent, and we selected E5-Large for the main analysis on this basis. The trained neural network weights for all models described in this paper are publicly available, along with code to reproduce all results, at [URL to be added upon publication].

\end{document}